\documentclass[letterpaper, 10 pt, conference]{ieeeconf}  

\IEEEoverridecommandlockouts                              

\usepackage{amsmath} 
\usepackage{amssymb}  
\usepackage{algorithm}
\usepackage{algpseudocode}
\usepackage{cite}
\usepackage{graphicx}
\usepackage{caption}
\usepackage{subcaption}
\usepackage{tabularx}
\usepackage{url}

\title{\LARGE \bf
Residual Physics Learning and System Identification for Sim-to-real Transfer of Policies on Buoyancy Assisted Legged Robots 
}

\author{Nitish Sontakke$^{1}$ Hosik Chae$^{2}$ Sangjoon Lee$^{3}$ Tianle Huang$^{1}$ Dennis W. Hong$^{2}$ Sehoon Ha$^{1}$
\thanks{$^{1}$School of Interactive Computing, Georgia Institute of Technology, Atlanta, GA, 30308, USA. {\tt \small nitishsontakke@gatech.edu, thuang325@gatech.edu sehoonha@gatech.edu}}%
\thanks{$^{2}$Department of Mechanical and Aerospace Engineering, University of California, Los Angeles (UCLA), Los Angeles, CA, 90095, USA. {\tt\small hosikchae@ucla.edu, dennishong@ucla.edu}}%
\thanks{$^{3}$Department of Computer Science, 
University of California, Los Angeles (UCLA), Los Angeles, CA, 90095, USA. {\tt\small sangjoonlee@cs.ucla.edu}}%
}

\usepackage{xcolor}
\newcommand{\cmt}[1]{}

\long\def\ignorethis#1{}

\newcommand{\etal}{{\em{et~al.}\ }}

\newcommand{\figref}[1]{Fig.~\ref{fig:#1}}

\newcommand{\vc}[1]{\ensuremath{\mathbf{#1}}}

\newcommand{\pctab}{\hspace{0.2in}}

\begin{document}

\maketitle
\thispagestyle{empty}
\pagestyle{empty}

\begin{abstract}

The light and soft characteristics of Buoyancy Assisted Lightweight Legged Unit (BALLU) robots have a great potential to provide intrinsically safe interactions in environments involving humans, unlike many heavy and rigid robots. However, their unique and sensitive dynamics impose challenges to obtaining robust control policies in the real world. In this work, we demonstrate robust sim-to-real transfer of control policies on the BALLU robots via system identification and our novel residual physics learning method, Environment Mimic (EnvMimic). First, we model the nonlinear dynamics of the actuators by collecting hardware data and optimizing the simulation parameters. Rather than relying on standard supervised learning formulations, we utilize deep reinforcement learning to train an external force policy to match real-world trajectories, which enables us to model residual physics with greater fidelity. We analyze the improved simulation fidelity by comparing the simulation trajectories against the real-world ones. We finally demonstrate that the improved simulator allows us to learn better walking and turning policies that can be successfully deployed on the hardware of BALLU.

\end{abstract}

\section{Introduction}
\begin{figure}
    \centering
    \includegraphics[width=\linewidth]{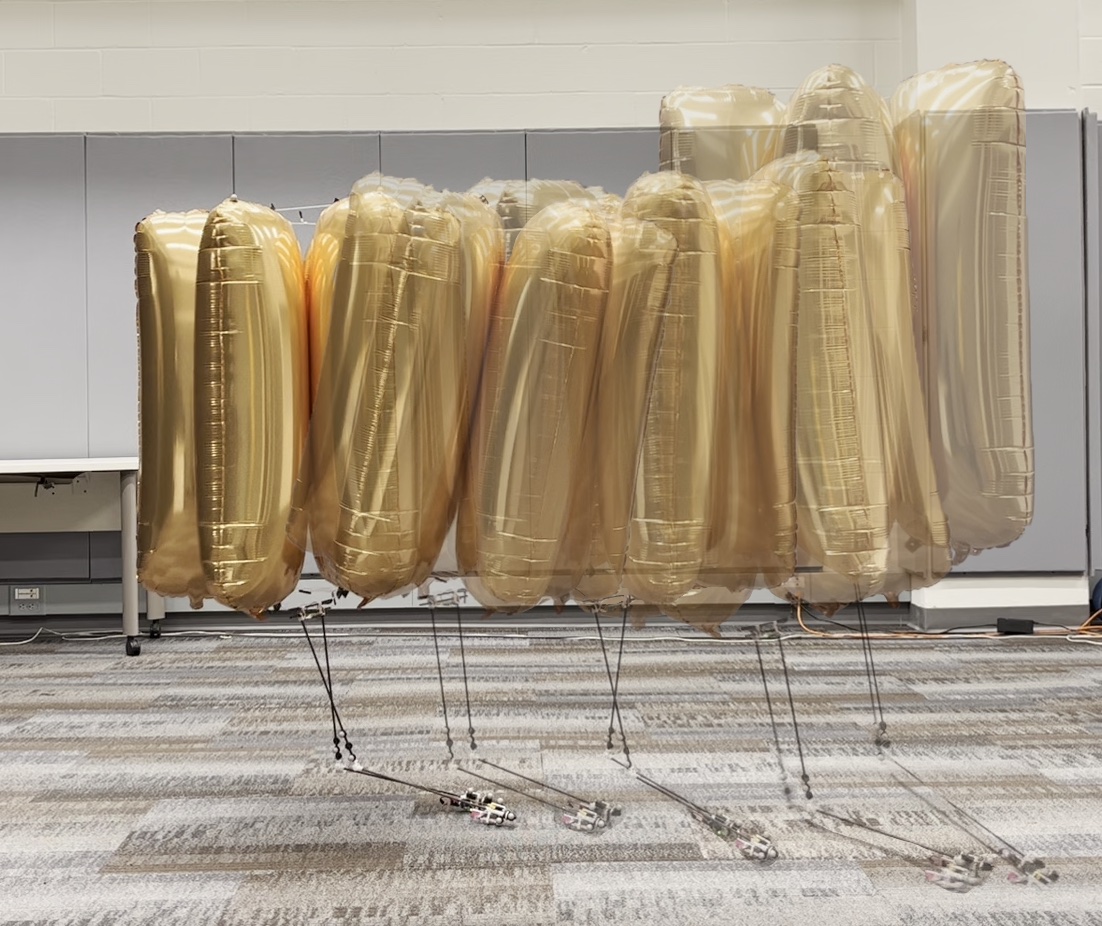}
    \caption{An image of the successful policy for the forward walking task, which is trained in the improved simulation using our method.}
    \label{fig:teaser}
\end{figure}

Buoyancy-assisted or balloon-based robots \cite{chae2021ballu2,balloonhumanoid2015,inflatablearm2017} have great potential to offer fundamental safety in human environments. Traditional mobile robots while being able to execute a variety of tasks, tend to be rigid and heavy and may cause serious damage to their surroundings or themselves in case of control or perception errors. On the other hand, buoyancy-assisted robots (BARs) \cite{williams2021introduction} are typically designed to be lightweight, compact, and intrinsically safe. Therefore, they can be used for various applications that require close human-robot interaction, such as education, entertainment, and healthcare. For instance, Chae \etal~\cite{chae2021ballu2} present Buoyancy Assisted Lightweight Legged Unit (BALLU), which is a balloon-based robot with two legs (\figref{teaser}), and showcase that it can be deployed to various indoor and outdoor environments without any safety concerns.

However, it is not straightforward to control BARs due to their unique, non-linear, and sensitive dynamics. One popular approach for robot control is model-predictive control (MPC)~\cite{shen2021novel} which plans future trajectories via models and minimizes the provided cost function. The complex dynamics of BARs, however, prevent us from developing concise and effective models and therefore rule out MPC as a control method. In contrast, deep reinforcement learning (deep RL) offers an automated approach to training a control policy from a simple reward function without robot-specific models. On the flip side, policies trained with deep RL often experience severe performance degradation when deployed to the robot due to the difference between the simulated and real-world environments, which is commonly known as the sim-to-real gap or the reality gap~\cite{jakobi1995noise}. In our experience, this gap is further compounded in the case of BALLU when we employ a vanilla rigid body simulator, such as PyBullet~\cite{coumans2021} or CoppeliaSim~\cite{coppeliaSim}, due to the unmodeled aerodynamics and low-fidelity actuators.

In this work, we mitigate the sim-to-real gap of the BALLU robot by identifying system parameters and modeling residual dynamics using a novel technique, \emph{EnvMimic}. First, we iteratively tune the actuator parameters in simulation based on the data collected from hardware experiments. This system identification allows us to better illustrate the nonlinear dynamics of BALLU's cable-driven actuation mechanism. Second, we learn the residual physics of the BALLU robot from the collected real-world trajectories to capture its complex aerodynamics that are difficult to model analytically. To this end, we propose a novel technique, Environment Mimic (EnvMimic), which learns to generate external forces to match the simulation and real-world trajectories via deep RL, which is different from common supervised learning formulations~\cite{ajay2018augmenting, golemo2018sim, bauersfeld2021neurobem, o2022neural}. This is similar to using pseudo-forces like centrifugal force or Coriolis force to explain the observed behavior. Our approach can also be viewed as an inside-out flipped version of the recent motion imitation frameworks~\cite{peng2018deepmimic, peng2021amp, peng2022ase, escontrela2022adversarial}, which learn internal controllers that enable the robot to imitate reference motions. In our case, we treat the real-world trajectories as a reference and learn an external residual force policy to imitate that behavior in simulation. We also observe that EnvMimic exhibits a robust generalization capability, even when we have a small number of trajectories.

We demonstrate that the proposed techniques can successfully reduce the sim-to-real gap of the BALLU robot. Firstly, we show that modeling the actuators and capturing the aerodynamics results in a significantly improved and qualitatively richer simulation. Our augmented simulator successfully illustrates asymmetric turning behaviors, which are observed on hardware but are not captured by the vanilla version or the simulator with supervised residual dynamics learning. We also demonstrate that we can improve the sim-to-real transfer performance of the policies for two tasks, walking and turning, on the hardware of the BALLU robot. 





\section{Related Work}


\subsection{Deep Reinforcement Learning}
Deep RL~\cite{pmlr-v48-mniha16,schulman2017proximal,haarnoja2018learning} has allowed researchers to make great strides in various fields of robotics, including navigation~\cite{bellemare2020autonomous,loquercio2021learning}, locomotion~\cite{miki2022learning,siekmann2021blind}, and manipulation \cite{peng2018sim,zeng2020tossingbot}. However, successfully deploying these controllers on hardware is still an active area of research~\cite{zhao2020sim, salvato2021crossing}, which is not straightforward due to the discrepancy between the simulation and the real world~\cite{jakobi1995noise}. One of the most common approaches is domain randomization ~\cite{tobin2017domain,peng2018sim,siekmann2020learning,siekmann2021sim,siekmann2021blind,li2021reinforcement}, which exposes an agent to a variety of dynamics during training. Additionally, employing extensions such as privileged learning~\cite{loquercio2021learning, miki2022learning} and adopting structured state~\cite{loquercio2021learning} and action space~\cite{kaufmann2022benchmark} representations has also enabled successful deployment of learned policies on hardware. On the other hand, researchers have developed frameworks to learn policies directly from real-world experience, which have been proven effective for both manipulators~\cite{zeng2020tossingbot} and legged robots~\cite{haarnoja2018learning,ha2020learning}. This paper discusses a sim-to-real transfer for the BALLU robot with highly sensitive dynamics inspired by these previous approaches. Drawing inspiration from these previous approaches, this paper discusses a sim-to-real transfer technique for the BALLU robot, which exhibits highly sensitive dynamics. Specifically, we train policies in simulation and enhance the sim-to-real transferability using real-world data.

\subsection{System Identification}
Our approach is also highly inspired by system identification that aims to identify model parameters from the collected experimental data. This is a well-studied problem that has been addressed by a variety of methods involving maximum likelihood estimation \cite{khosla1985parameter, gautier1988identification}, optimization-based strategies \cite{tan2016simulation, zhu2017fast, chebotar2019closing}, neural networks \cite{yu2017preparing, allevato2020tunenet, zhang2022zero} with iterative learning \cite{allevato2020iterative, du2021auto}, actuator dynamics identification~\cite{hwangbo2019learning, yu2019sim}, adversarial learning \cite{jiang2021simgan}, and learning residual physics \cite{ajay2018augmenting, golemo2018sim, zeng2020tossingbot, bauersfeld2021neurobem, o2022neural}. Combining system identification with other techniques, such as dynamics randomization, latency modeling, and noise injection \cite{tan2018sim, rodriguez2021deepwalk}, has also proven 
to be effective for successful sim-to-real transfer of learned policies. However, in our case, system identification even in combination with domain randomization, proves to be insufficient, necessitating the need for our residual dynamics learning framework, EnvMimic.


\subsection{Balloon-based Robots}
Balloon-based or buoyancy-assisted robots~\cite{balloonhumanoid2015,flyingactuator2018} have been investigated because of their intrinsic stability and low costs. Therefore, many researchers have investigated them in various applications, including roof cleaning~\cite{glassballoon2002}, planet exploration~\cite{nayar2019balloon}, disaster investigation~\cite{ballooncablerobot2005, yamada2018gerwalk, takeichi2017development}, social interactions~\cite{balloonfriend2015}, security~\cite{balloonsysid2022}, and many more. However, control of these balloon robots is not straightforward due to their sensitive non-linear dynamics. One common approach is to develop model-based controllers~\cite{fullstatebuav2021,finballoon2021,balloonsysid2022,inflatablearm2017}, often with system identification. This paper discusses the control of Balloon-based legged robots proposed by Chae \etal~\cite{chae2021ballu2} by leveraging deep reinforcement learning and residual physics.

\section{Sim-to-real of BALLU}
In this section, we will describe our techniques for reducing the `reality gap'~\cite{jakobi1995noise} of the BALLU robot~\cite{chae2021ballu2}. We approach this challenging problem by combining traditional system identification and deep residual dynamics learning. First, we improve the simulation model of cable-driven actuation by identifying non-linear relationships between motor and joint angles. Next, we use the captured real-world trajectories to model the residual dynamics of the BALLU robot, which arise from various sources such as aerodynamics, joint slackness, and inertial parameter mismatch. Our key invention is to use deep RL for building a residual dynamics model instead of the common choice of supervised learning, which offers effective generalization over a small number of trajectories.


\subsection{Background: BALLU robot}
\begin{figure}
    \centering
    \begin{subfigure}{0.49\linewidth}
    \includegraphics[width=\linewidth]{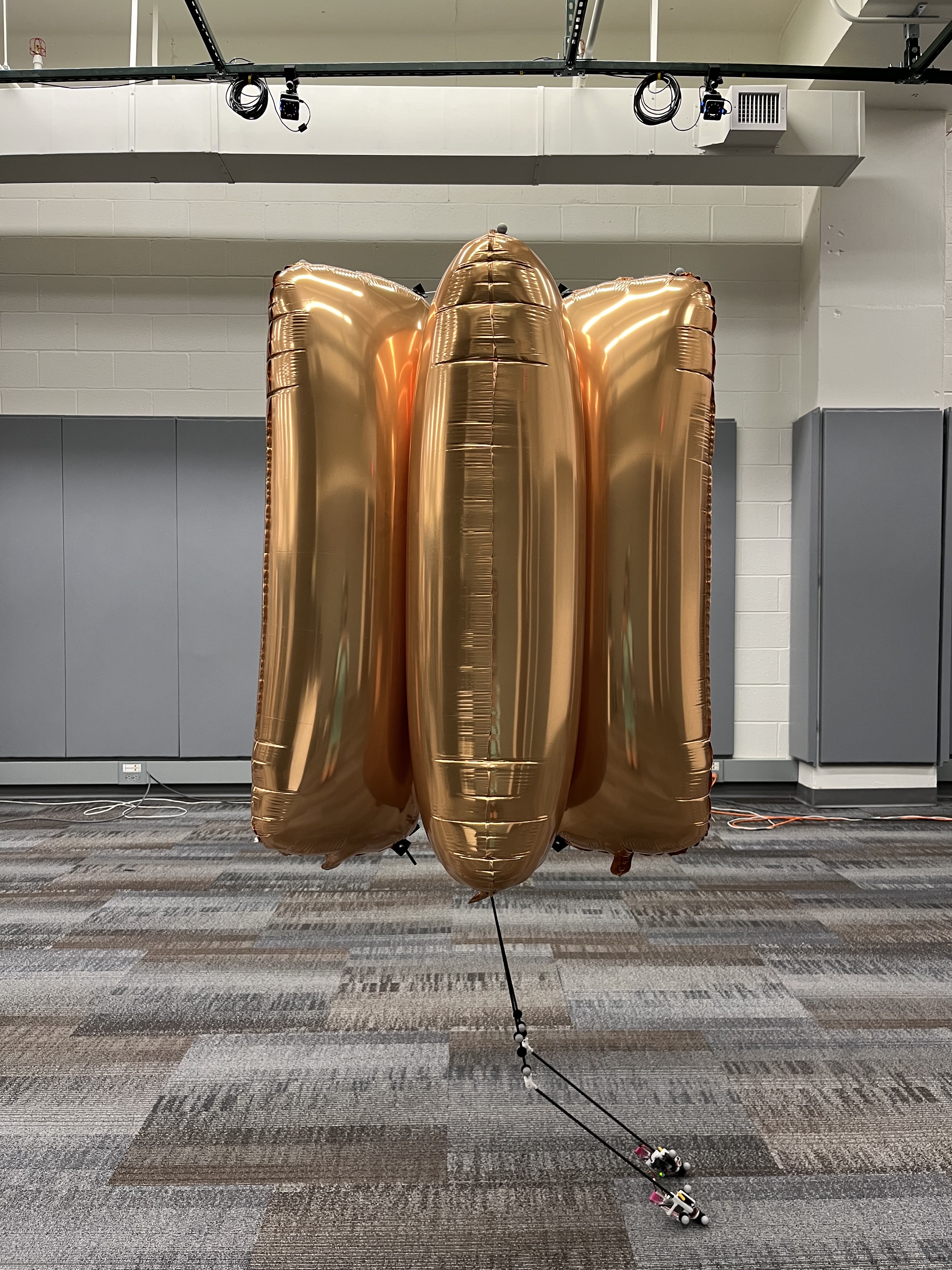}
    \label{fig:ballu_full_side}
    \caption{Side view}
    \end{subfigure}
    \hfill
    \begin{subfigure}{0.49\linewidth}
    \includegraphics[width=\linewidth]{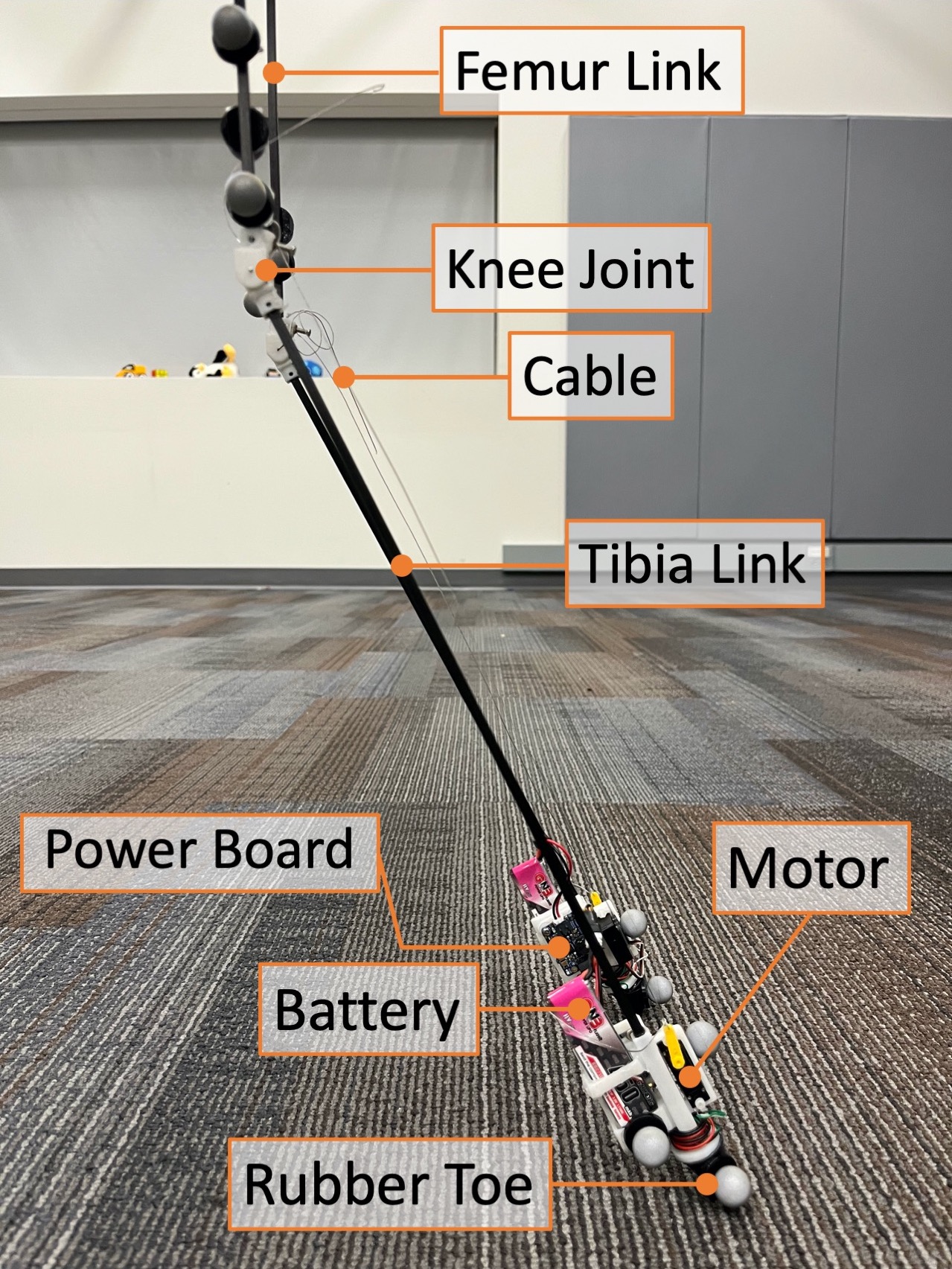}
    \label{fig:ballu_knee}
    \caption{Leg details}
    \end{subfigure}
    \caption{Illustration of our research platform, BALLU (Buoyancy Assisted Lightweight Legged Unit) with two passive hip joints and two active knee joints.}
    \label{fig:ballu}
\end{figure}
BALLU (Buoyancy Assisted Lightweight Legged Unit) is a novel buoyancy-assisted bipedal robot with six helium balloons, which provide enough buoyancy to counteract the gravitational force. BALLU's base is connected to helium balloons and houses a Raspberry Pi Zero W board for computing. The robot has two passive hip joints and two active knee joints, which are actuated by two Dymond D47 servo motors at the feet via cables. The overview of the robot is illustrated in \figref{ballu}. For more details, please refer to the original paper by Chae \etal colleague~\cite{chae2021ballu2}.

Due to its unique dynamics, model-free reinforcement learning can be a promising approach for developing effective controllers for BALLU without having to rely on prior knowledge or domain expertise. However, we need to mitigate the large sim-to-real gap first, which is induced by significant drag force effects and low-fidelity hardware.

\subsection{System Identification} \label{sec:system_id}
One main source of the sim-to-real gap is its cable-driven actuation mechanism. In the simulation, servo motor commands and knee joint angles maintain an ideal relationship. In reality, they are affected by friction, torque saturation, and unmodeled cable dynamics, which make the actuator dynamics noisy and nonlinear. 
Therefore, we first perform system identification to better capture this nonlinear relationship from real-world data using optimization.

Our free variables $\vc{p}$ include knee spring parameters, motor gains, default motor angles, and default knee joint angles in simulation, which are sufficient to model various nonlinear relationships. As a result, we have eight free variables subject to optimization. 

Our objective function is to minimize the discrepancy of all four joint angles (left and right, motor arm and knee) between simulation and hardware. We sample $20$ actuation commands that constitute $\mathcal{A}$ that are uniformly distributed over the range [$0, 1$], which corresponds to motor arm angles in the range [$0^\circ$, $90^\circ$], and measure {knee and motor} joint angles in simulation and on hardware. Then we fit polynomial curves for all the joints and compute the directed Hausdorff distance between the corresponding curves. We use the {L-BFGS-B} algorithm and optimize the parameters until convergence. The entire process is summarized in Algorithm 1.



\begin{algorithm}
\caption{System Identification of Cable-driven Actuation}
\label{sysID}
\begin{algorithmic}[1]
\State{\textbf{Input:} the initial parameters $\vc{p}_0$}
\State{\textbf{Input:} a set of pre-defined actions $\mathcal{A}$}
\State{Measure joint angles on hardware for all actions $\mathcal{A}$}
\State{Fit polynomial curves $\overline{C}_1$, $\overline{C}_2$, $\overline{C}_3$, and $\overline{C}_4$}
\State{$\vc{p} \leftarrow \vc{p}_0$}
\While{not converged}
    \State{Update the simulation with $\vc{p}$}
    \State{Measure joint angles for all actions $\mathcal{A}$}
    \State{Fit polynomial curves $C_1$, $C_2$, $C_3$, and $C_4$}
    \State{$\epsilon \leftarrow$ directed Hausdorff distance between $C_i$ and $\overline{C}_i$}
    \State{Optimize $\vc{p}$ using {L-BFGS-B}}
\EndWhile
\end{algorithmic}
\end{algorithm}

\subsection{Residual Dynamics Learning via Reinforcement Learning} \label{sec:envmimic}
Our next step is to model the residual dynamics of BALLU. Previous methods for learning residual dynamics have employed supervised learning~\cite{golemo2018sim, ajay2018augmenting, bauersfeld2021neurobem, o2022neural} or a combination of self-supervision with deep RL as part of the learning pipeline~\cite{zeng2020tossingbot}. However, off-the-shelf supervised learning, in addition to requiring a large number of real-world trajectories, is plagued by limited exploration. Even a small perturbation to the states the policy has observed during training can cause it to diverge during test time. Moreover, we observe that stochasticity in the real world often leads to multiple different state trajectories arising from the same state even when we apply the same actions. 
The framework of deep RL lends itself naturally to addressing these issues by augmenting the data with simulated trajectories, making it a suitable choice for this problem.


Our key insight is to augment the original simulation framework using a learned residual aerodynamics policy. This policy allows us to capture the complex interaction between BALLU and its environment in greater detail. We will demonstrate in the next section that learning locomotion behaviors with this aerodynamics policy in the loop translate to better transfer of our simulation policies to hardware compared to traditional techniques like domain randomization.

Therefore, we design a framework to learn a policy that generates proper \emph{external} perturbation forces that can match the simulation behavior to the ground-truth trajectory collected from the hardware. We draw inspiration from motion imitation methods~\cite{peng2018deepmimic, peng2021amp, peng2022ase, escontrela2022adversarial}, which have demonstrated impressive results for learning dynamics controllers to track reference motions. The fundamental difference is that we learn a policy for external perturbations, while other motion imitation works aim to learn an internal control policy for the robot's actuators. In our experience, this deep RL approach allows us to model robust residual dynamics from a limited set of real-world trajectories compared to supervised learning.

\noindent \textbf{Data Collection.} The first step is to create a set of reference trajectories. We train several locomotion policies in the vanilla simulation and record their action trajectories. Next, we use the recorded actions as open loop control on hardware to collect multiple state trajectories. We use a motion capture system to obtain observation data due to the lack of onboard sensors on BALLU that can estimate its global position and orientation. We note that hand-designed action trajectories may work well for this step. 

\noindent \textbf{MDP Formulation.}
Once we have the reference dataset, we can cast learning the residual aerodynamics policy as a motion imitation problem using a Markov Decision Process (MDP). The state space consists of the balloon's position, velocity, orientation, the position and velocity of the base, and the position and velocity of the feet, at the current and last two time steps. The action space is three-dimensional and consists of x, y, and z forces that are applied to the center of mass of the balloon. The forces are in the range of $[-1, 1]$ N. The reward function is a combination of position and orientation terms and is defined as follows:
\begin{align*}
    r_t = w^{pos} r^{pos}_t + w^{orn} r^{orn}_t
\end{align*}
where the position and orientation terms respectively are computed as follows:
\begin{align*}
    &r^{pos} = \text{exp}\left[-10 \left( || \hat{\vc{p}}_t - \vc{p}_t||^2 \right) \right] \\
    &r^{orn} = \text{exp}\left[-2 \left( || \hat{\vc{r}}_t - \vc{r}_t||_W^2 \right) \right],
\end{align*}
where $\hat{\vc{p}}_t$, $\vc{p}_t$, $\hat{\vc{r}}_t$, and $\vc{r}_t$ are the desired position, the actual position, the desired orientation, and the actual orientation of the balloons, respectively. The position reward $r^{pos}_t$ encourages the simulated model's balloon to track the reference balloon position as closely as possible while the orientation reward $r^{orn}_t$ encourages it to track the reference balloon orientation. We use the Euler angle representation for orientation, which demonstrates better performance than the quaternion representation. For all experiments, we set $w^{pos} = 0.7$, $w^{orn} = 0.3$, and $W = diag(0.2, 0.4, 0.4)$.

\noindent \textbf{Training.}
We train the residual dynamics policy using Proximal Policy Optimization~\cite{schulman2017proximal}. We use a compact network consisting of two layers with $64$ neurons each. Similar to Peng \etal~\cite{peng2018deepmimic}, we also randomize the initial state for each rollout by sampling a state uniformly at random from the selected reference trajectory. This leads to the policy being exposed to a wider initial state distribution and improves robustness, especially when transferring to hardware.

\subsection{Policy Training with Improved Simulation} \label{sec:policy}
Once we improve the simulation using system identification and residual dynamics learning, we can retrain a deep RL policy for better sim-to-real transfer. Once again, we formulate the problem using a Markov Decision Process framework. The state space consists of the balloon's position, velocity, orientation, the position and velocity of the base, and the position and velocity of the feet, all measured at the current time step.
Our actions are two actuator commands, which will change the joint angles based on the identified nonlinear relationship in the previous section. 

We learn two policies - one for forward walking and one for turning left. For the forward walking task, our reward function is $x_{vel}$, whereas for turning left, it is $y_{vel}$.

\section{Experiments and Results}
We design simulation and hardware results to answer the following two research questions. 
\begin{itemize}
\item Can we improve the fidelity of the vanilla simulator using actuator identification and residual dynamics learning?
\item Can we improve the performance on hardware by reducing the sim-to-real gap?
\end{itemize}

\subsection{Experimental Setup}
We conduct all the simulation experiments in PyBullet~\cite{coumans2016pybullet}, an open-source physics-based simulator. We use the stable baselines~\cite{raffin2021stable} implementation of Proximal Policy Optimization~\cite{schulman2017proximal} to learn the residual dynamics (Section~\ref{sec:envmimic}) and the policy in the improved simulation (Section~\ref{sec:policy}). We use the BALLU platform~\cite{chae2021ballu2} for hardware experiments while capturing all the data using a Vicon motion capture system~\cite{vicon}. 



\subsection{Improved Simulation Fidelity}
This section illustrates the process for improving the simulation's fidelity. We first highlight the importance of actuator system identification in Section~\ref{sec:actuator_results} and show the learned residual dynamics using our EnvMimic technique in Section~\ref{sec:envmimic_results}.

\subsubsection{Actuator System Identification} \label{sec:actuator_results}

\begin{figure}
    \centering
    \includegraphics[width=\linewidth]{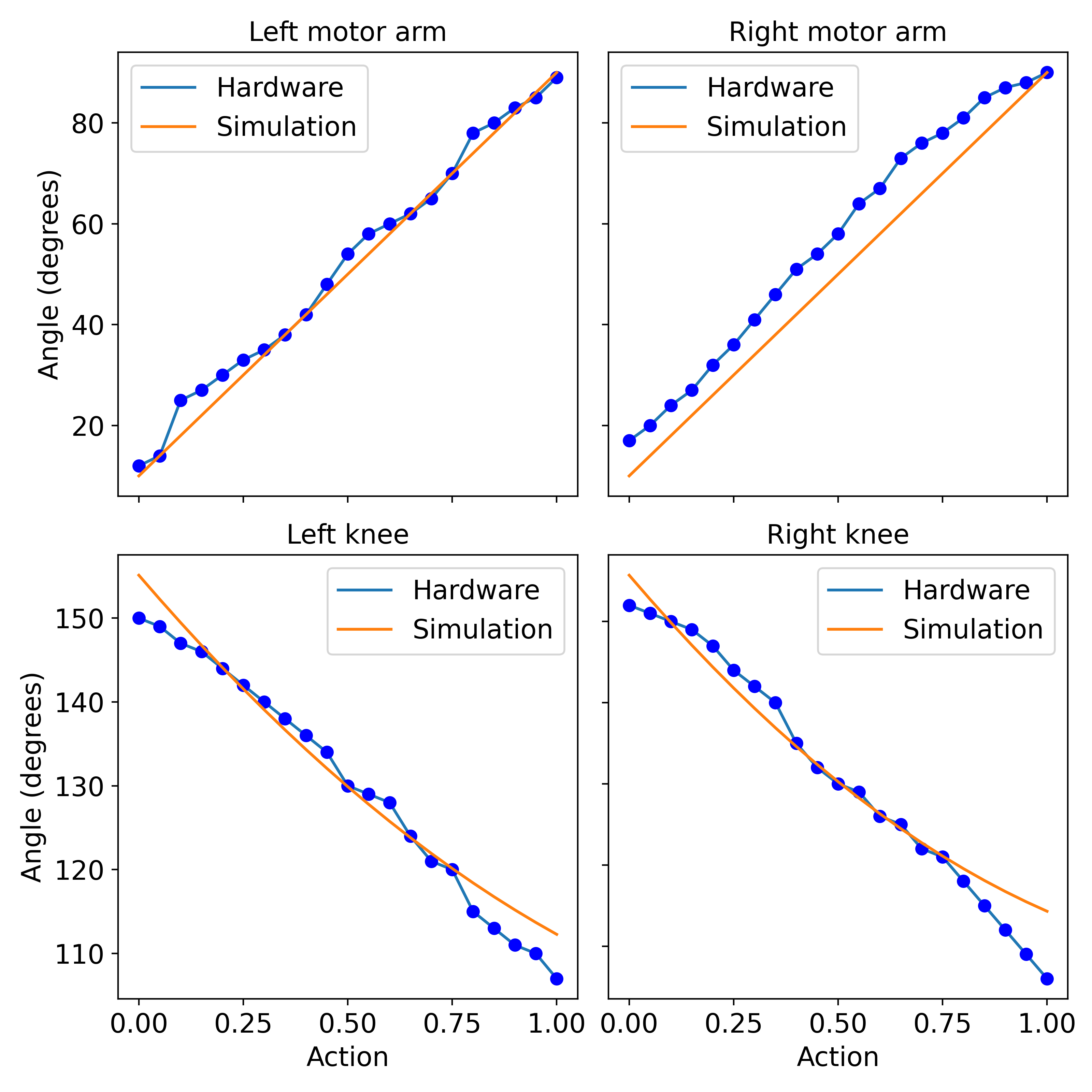}
    \caption{Identified non-linear, asymmetric relationships of cable-driven mechanisms.}
    \label{fig:actuator_sys_ID}
\end{figure}

We collect the data and identify the system parameters, such as spring parameters, motor gains, and default joint angles, as described in Section~\ref{sec:system_id}. The identified relationships between the motor commands and joint angles are illustrated in Fig.~\ref{fig:actuator_sys_ID}. As shown, the identified relationships exhibit highly nonlinear behaviors compared to the simple idealized curves in simulation, which are essential to model the dynamics of the BALLU robot.

\begin{figure}
    \centering
    \begin{subfigure}{0.4925\linewidth}
    \includegraphics[width=\linewidth]{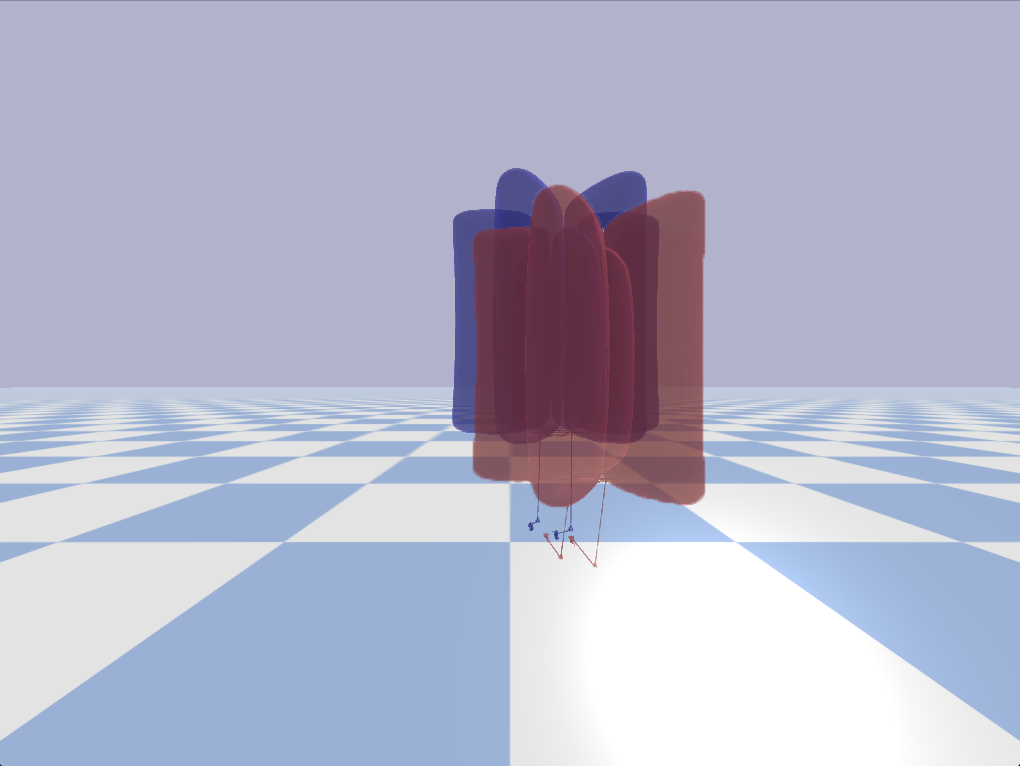}
    \label{fig:sys_ID_overlay_front}
    \end{subfigure}
    \hfill
    \begin{subfigure}{0.4925\linewidth}
    \includegraphics[width=\linewidth]{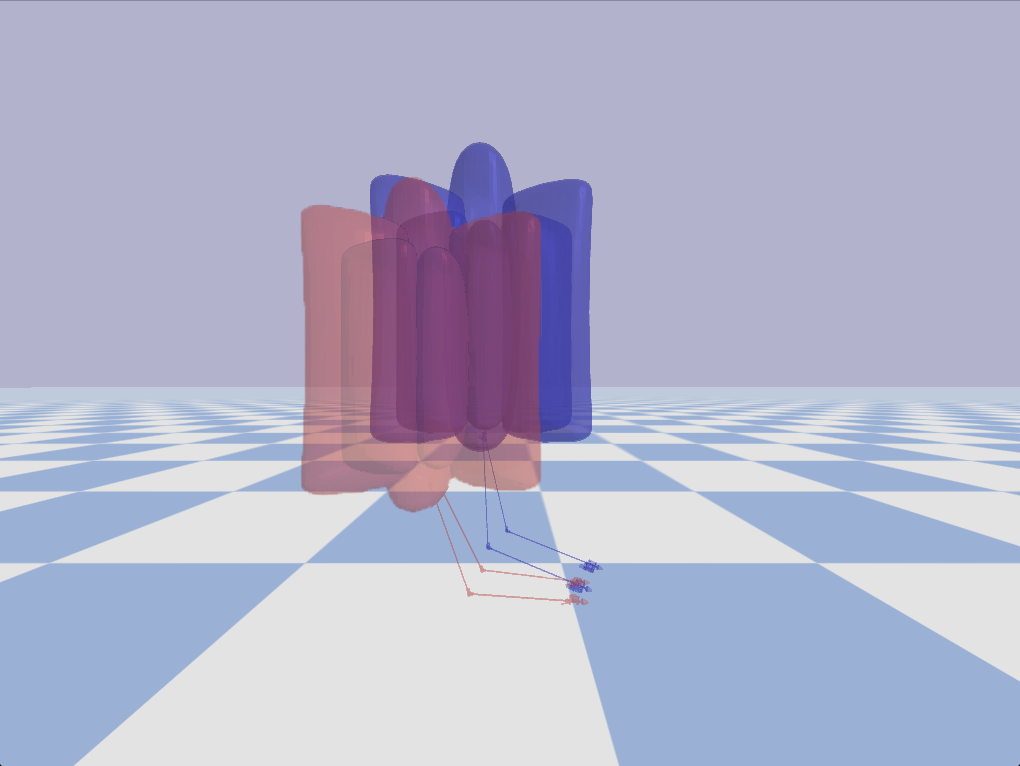}
    \label{fig:sys_ID_overlay_side}
    \end{subfigure}
    \caption{Illustration of System Identification Results. We execute the same action trajectory and compare the final state of identified dynamics (blue) to that of the naive simulation (red), which is significantly different.}
    \label{fig:sys_ID_overlay}
\end{figure}

We highlight the importance of system identification by comparing trajectories in simulation. We run the same action sequences of periodic bang-bang control signals with and without system identification and compare the final states in Fig.~\ref{fig:sys_ID_overlay}. The two generated trajectories show a significant difference in terms of the final COM positions ($0.23$~m difference) and the joint angles ($10.15^\circ$ difference, average of all joints). 

\subsubsection{EnvMimic: Residual Dynamics Learning} \label{sec:envmimic_results}
Next, we examine the results of residual dynamics learning using the proposed EnvMimic technique in Section~\ref{sec:envmimic}. We hypothesize that EnvMimic can learn compelling residual dynamics from a few trajectories, unlike data-hungry supervised learning approaches. We compare the x-y-yaw trajectories in four different environments: (1) a vanilla simulation, (2) a simulation with the residual dynamics learned with supervised learning, (3) a simulation with the residual dynamics learned with EnvMimic (ours), and (4) the ground-truth trajectory on hardware. For supervised learning, we use a neural work with two hidden layers of size [64, 64].
For all the trajectories, we use the same action sequences that are generated by the initial policy in a vanilla simulation. Also, please note that the testing ground-truth trajectory is unseen during training. 

\begin{figure}
    \centering
    \includegraphics[width=\linewidth]{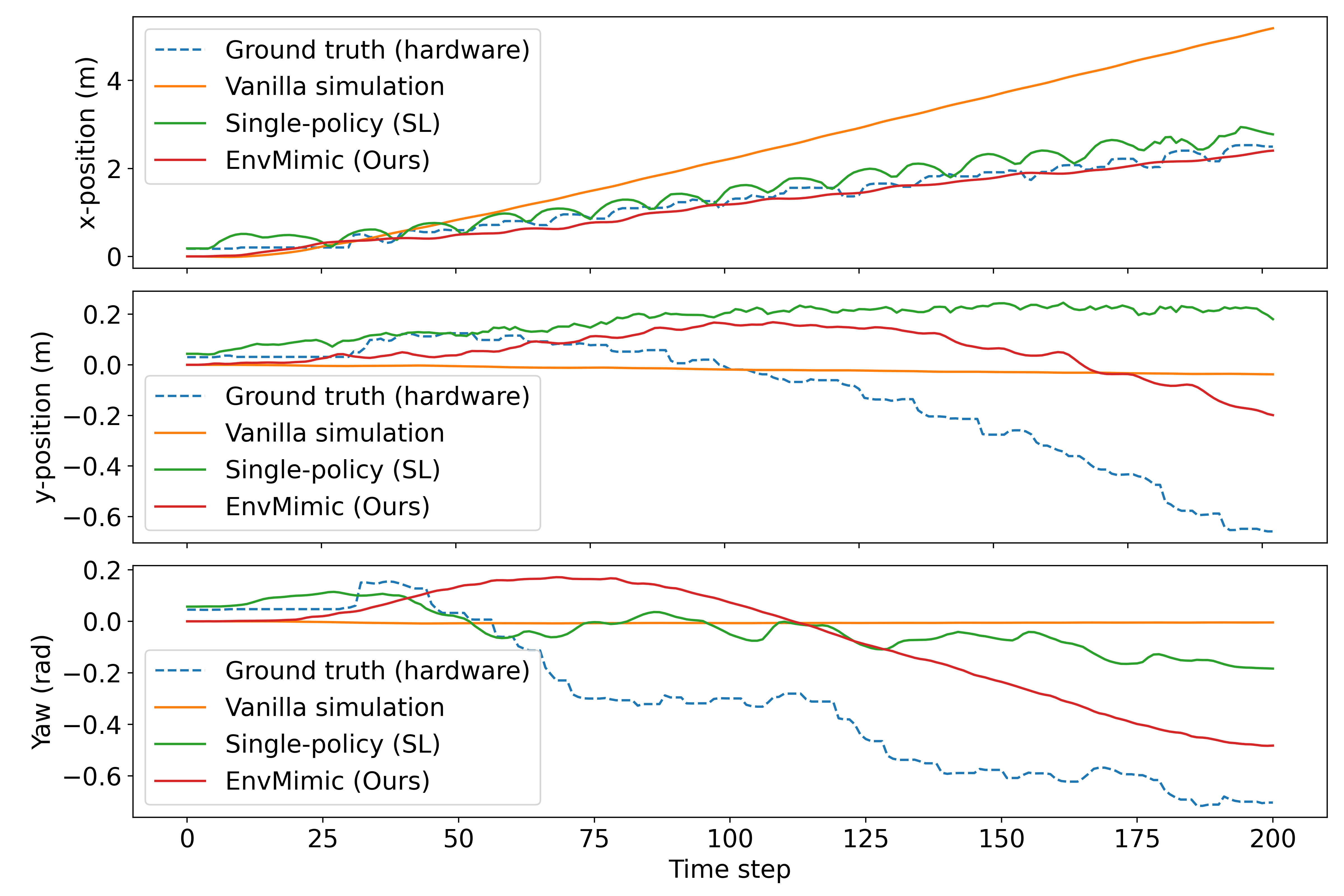}
    \caption{Comparison of simulation trajectories to ground truth hardware data for forward walking. Our method, EnvMimic, shows the best tracking performance, particularly in terms of the yaw angle. Note that the ground truth trajectory shown is out-of-distribution.}
    \label{fig:envmimic}
\end{figure}

The trajectories are compared in Fig.~\ref{fig:envmimic}.
Clearly, our EnvMimic offers much-improved tracking performance compared to the vanilla simulation that fails to capture the noticeable yaw orientation changes due to the stochasticity of the hardware experiments. In our experience, the trajectory generated with supervised residual dynamics learning tends to turn less and remains on the positive Y side. We hypothesize that the robustness of our RL-based residual dynamics might be obtained due to the mix-use of real-world and simulation trajectories. On the other hand, the supervised learning baseline is trained on the pre-collected hardware trajectories without any data augmentation. We believe the comparison of SL or RL-based approaches on a wider range of scenarios will be an interesting future research direction. Please refer to the supplemental video for qualitative comparisons, highlighting the obvious benefit of our EnvMimic-based residual dynamics learning. 

\subsection{Improved Sim-to-real Transfer}
\begin{table}
    \centering
    \begin{tabularx}{\linewidth}{ 
    | >{\centering\arraybackslash}X
    | >{\centering\arraybackslash}X
    | >{\centering\arraybackslash}X
    | >{\centering\arraybackslash}X|}
    \hline
        \textbf{Experiment} &$(CoM^{sim}_t - CoM^{hw}_t)/T$(m) & \textbf{x-distance traveled (m)} & Total distance traveled (m) \\
        \hline
        Vanilla + sys ID & $2.01$ & $0.27$ & $1.24$ \\
        \hline
        Vanilla + DR & $1.24$ & $0.32$ & $0.60$ \\
        \hline
        Vanilla + sys ID + DR & $1.07$ & $0.56$ & $0.94$ \\
        \hline
        EnvMimic (Ours) & $\mathbf{0.90}$ & $\mathbf{1.12}$ & $\mathbf{1.28}$ \\
    \hline
    \end{tabularx}
    \caption{Sim-to-real Comparison for Forward Walking.}
    \label{table:fwd_walk}
\end{table}

\begin{table}
    \centering
    \begin{tabularx}{\linewidth}{ 
    | >{\centering\arraybackslash}X
    | >{\centering\arraybackslash}X
    | >{\centering\arraybackslash}X
    | >{\centering\arraybackslash}X|}
    \hline
        \textbf{Experiment} & $\alpha^{sim}_T - \alpha^{hw}_T$ & \textbf{y-distance traveled (m)} & $\Delta \alpha $ (hardware)\\
        \hline
        Vanilla + sys ID & $16.41^{\circ}$ & $0.20$ & $16.72^{\circ}$\\
        \hline
        Vanilla + DR & $38.18^{\circ}$ & $-0.04$ & $-4.28^{\circ}$\\
        \hline
        Vanilla + sys ID + DR & $-8.50^{\circ}$ & $0.11$ & $\mathbf{39.77^{\circ}}$ \\
        \hline
        EnvMimic (Ours) & $\mathbf{3.50^{\circ}}$ & $\mathbf{0.29}$ & $36.42^{\circ}$\\
    \hline
    \end{tabularx}
    \caption{Sim-to-real Comparison for Turning Left.}
    \label{table:left_turn}
\end{table}

\begin{figure*}
    \centering
    \setlength{\tabcolsep}{1pt}
    \renewcommand{\arraystretch}{0.7}
    \begin{tabular}{c c c c c}
    \includegraphics[width=0.195\textwidth]{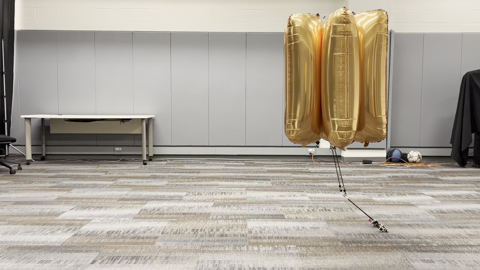} &
    \includegraphics[width=0.195\textwidth]{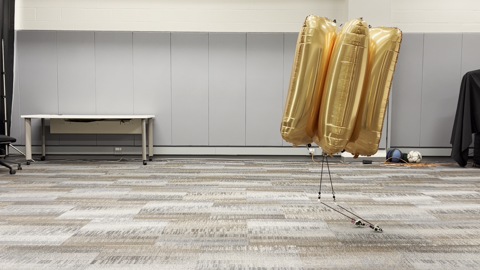} &
    \includegraphics[width=0.195\textwidth]{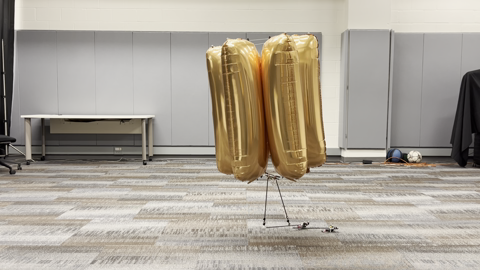} &
    \includegraphics[width=0.195\textwidth]{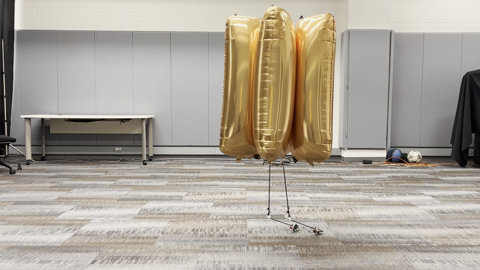} &
    \includegraphics[width=0.195\textwidth]{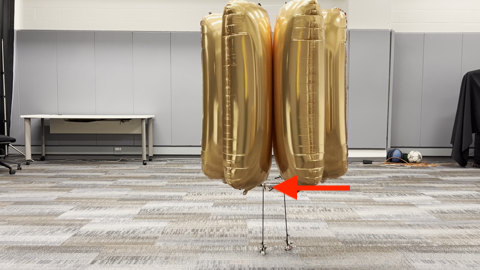} \\
    \includegraphics[width=0.195\textwidth]{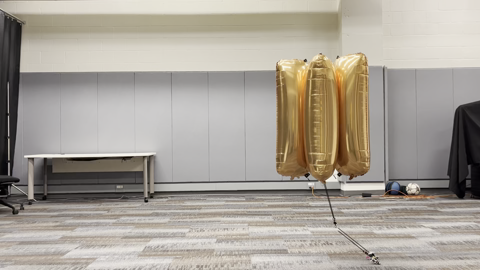} &
    \includegraphics[width=0.195\textwidth]{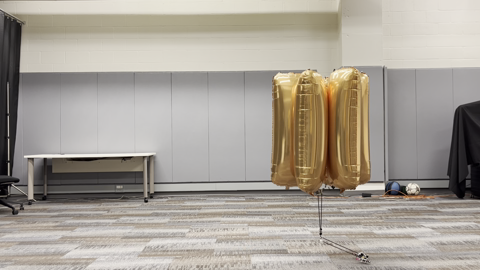} &
    \includegraphics[width=0.195\textwidth]{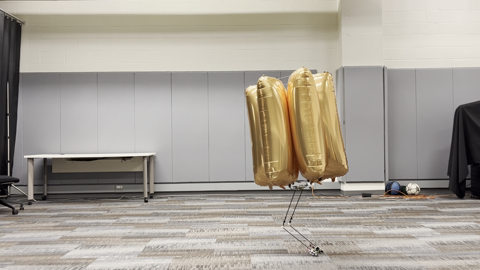} &
    \includegraphics[width=0.195\textwidth]{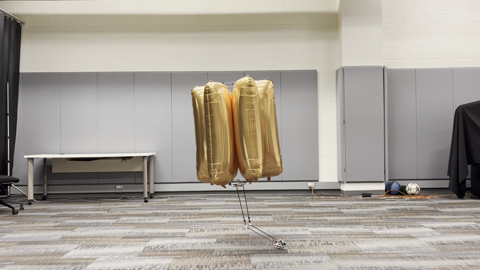} &
    \includegraphics[width=0.195\textwidth]{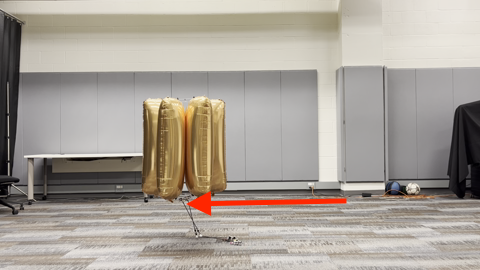} \\
    $t=0$s & $t=5$s & $t=11$s & $t=16$s & $t=24$s\\
    \end{tabular}
    
    \caption{Comparison of the learned forward walking policies without (\textbf{top}) and with (\textbf{bottom}, ours) the proposed residual dynamics learning. Both policies are also trained with domain randomization and actuator system identification. Please note that the baseline (\textbf{top}) shows a significant turning, while ours (\textbf{bottom}) can walk double the distance of the baseline.
    }
    \label{fig:hardware_forward}	
\end{figure*}

\begin{figure*}
    \centering
    \setlength{\tabcolsep}{1pt}
    \renewcommand{\arraystretch}{0.7}
    \begin{tabular}{c c c c c}
    \includegraphics[width=0.195\textwidth]{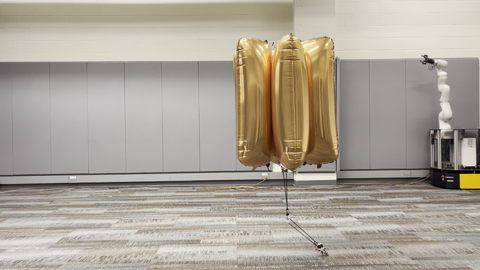} &
    \includegraphics[width=0.195\textwidth]{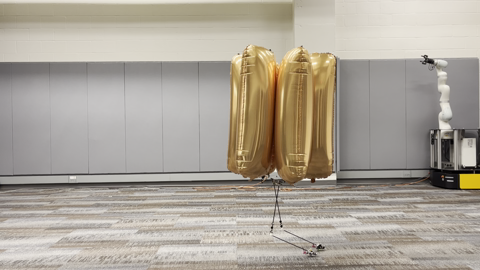} &
    \includegraphics[width=0.195\textwidth]{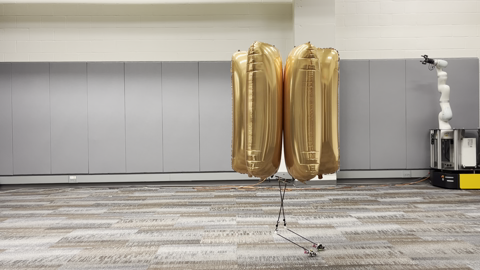} &
    \includegraphics[width=0.195\textwidth]{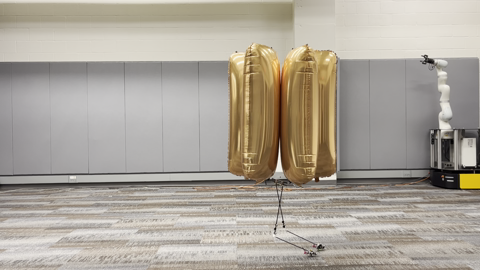} &
    \includegraphics[width=0.195\textwidth]{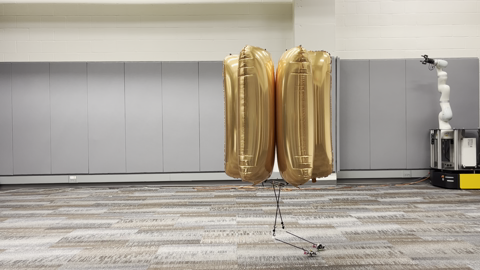} \\
    \includegraphics[width=0.195\textwidth]{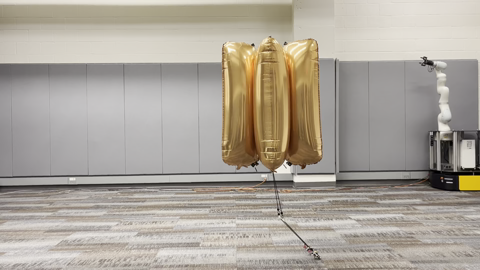} &
    \includegraphics[width=0.195\textwidth]{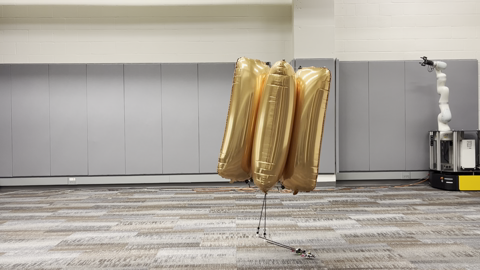} &
    \includegraphics[width=0.195\textwidth]{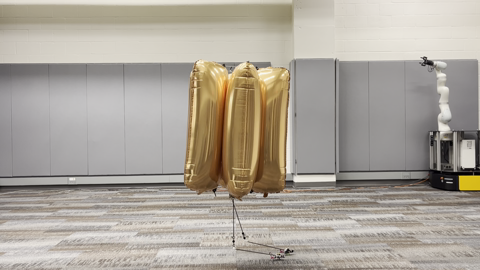} &
    \includegraphics[width=0.195\textwidth]{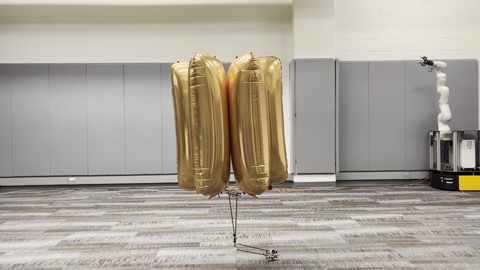} &
    \includegraphics[width=0.195\textwidth]{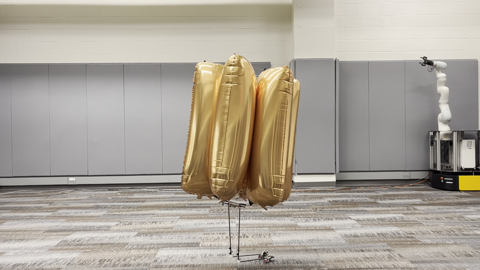} \\
    $t=0$s & $t=4$s & $t=10$s & $t=18$s & $t=22$s\\
    \end{tabular}
    
    \caption{Comparison of the learned turning left policies without (\textbf{top}) and with (\textbf{bottom}, ours) the proposed residual dynamics learning. Both policies are also trained with domain randomization and actuator system identification. Both policies are able to achieve a similar change in yaw angle ($\Delta \alpha$) over the entire episode, but the baseline (\textbf{top}) takes a single step and only turns in place, rotating over the battery cover. We also observe that our method (\textbf{bottom}) covers more than twice the distance in the desired y-direction.
    }
    \label{fig:hardware_left}	
\end{figure*}

To complete the story, we investigate whether we can improve the sim-to-real transfer of policies using augmented simulation. We first train (1) a policy in the improved simulation with the system identification, learned residual dynamics, and domain randomization~\cite{tobin2017domain} (ours) and compare the performance with the selected baseline policies learned in the following settings: (2) a simulation only with system identification (Vanilla + sys ID), (3) a simulation only with domain randomization (Vanilla + DR), and (4) a simulation with both system identification and domain randomization (Vanilla + sys ID + DR). For domain randomization, we randomly sample parameters for friction and initial states. We decided not to randomize masses or the buoyancy coefficient due to the sensitivity of the policy to these parameters. We evaluate these simulation-learned policies on the hardware and measure their performance. 

For the forward walking task, the learned policy with our augmented simulation is the only one which can walk forward while the other baselines turn left significantly. Therefore, its traveled distance in the forward (x) direction is $1.12$~m, which is significantly larger than $0.27$~m, $0.32$~m, and $0.56$~m of the others. For the turning task, our approach trains the effective policy that travels the most distance in the left (y) direction, $0.29$~m, which is our objective function. On the other hand, the other policies cover a shorter distance: $0.20$~m, $-0.04$~m, and $0.11$m. We note that the change in yaw angle ($\Delta \alpha$) is slightly larger in the case of the baseline with system identification and domain randomization compared to our method. This is an artifact that arises from the policy taking a single step and rotating on the battery cover. This is evident from the y-distance traveled and can be observed clearly in the qualitative results in Figure \ref{fig:hardware_left} and the supplemental video. 

For both tasks, our augmented simulation also exhibits the least sim-to-real errors, which are defined as the average center of mass (CoM) error and final yaw angle ($\alpha$) error between simulation and hardware. The performance is summarized in Table~\ref{table:fwd_walk} and Table~\ref{table:left_turn}. Please refer to the supplemental video and Fig.~\ref{fig:hardware_forward} for qualitative comparison.  

\section{Conclusion and Discussion}
We present a learning-based method for the sim-to-real transfer of locomotion policies for the Buoyancy
Assisted Lightweight Legged Unit (BALLU) robot, which has unique and sensitive dynamics. To mitigate a large sim-to-real gap, we first identify nonlinear relationships between motor commands and joint angles. Then we develop a novel residual dynamics learning framework, \emph{EnvMimic}, which trains an external perturbation policy via deep reinforcement learning. Once we improve the simulation accuracy with the identified actuator parameters and the learned residual physics, we retrain a policy for better sim-to-real transfer. We demonstrate that using our methodology, we can train walking and turning policies that are successful on the hardware of the BALLU robot.

There exist several interesting future research directions we plan to investigate in the near future. In this work, we develop our residual dynamics model for each individual task, such as walking or turning, which limits generalization over other tasks. Therefore, it will be interesting if we collect a large dataset and train a general residual dynamics model using the proposed method. It will be possible to take some inspiration from the state-of-the-art motion imitation frameworks, which can track a large number of trajectories using a single policy~\cite{peng2021amp}. In addition, we also want to investigate various policy formulations. This paper assumes simple external forces to the center of the balloons to model aerodynamics, and it was sufficient for the locomotion tasks we tested on. However, we may need multiple forces or torques to model some sophisticated phenomena. Furthermore, the dynamics of the BALLU robot are also sensitive to time owing to the deflation of balloons. In the future, we want to introduce the concept of lifelong learning to model those gradual temporal changes.

Finally, we plan to evaluate the proposed residual dynamics learning approach, \emph{EnvMimic}, on different tasks and robotic platforms. While showing promising results, many experiments are limited to the selected walking and turning tasks and the specific hardware of BALLU. However, we believe the algorithm itself is agnostic to the problem formulation, and it has great potential to improve the sim-to-real transferability in general scenarios, even including drones and rigid robots. We intend to explore this topic further in future research.





\section*{Acknowledgement}

This work is supported by the National Science Foundation under Award \#2024768.


\bibliographystyle{ieeetr}
\bibliography{refs}

\end{document}